\documentclass{article}

\usepackage{microtype}
\usepackage{graphicx}
\usepackage{subfigure} 
\usepackage{booktabs}
\usepackage{hyperref}

\usepackage[T1]{fontenc}
\usepackage[utf8]{inputenc}
\usepackage{amsmath,amssymb,amsfonts}
\usepackage{bm}
\usepackage{xcolor}
\usepackage{tikz}
\usetikzlibrary{positioning,arrows.meta}

 \usepackage[accepted]{icml2021}

\icmltitlerunning{Field Converter: World-Grounded Player Pose Estimation}

\begin{document}


\twocolumn[
\icmltitle{
Field Converter: Geometry-Initialized Temporal Residual Refinement for
World-Grounded Player Pose Estimation from Soccer Broadcasts
}

\begin{icmlauthorlist}
    \icmlauthor{Simon Khan}{ibhgc,fff}
    \icmlauthor{Laurent Gajny}{ibhgc}
    \icmlauthor{Jennyfer Lecompte}{fff}
    \icmlauthor{Sébastien Laporte}{ibhgc}
\end{icmlauthorlist}

\icmlaffiliation{ibhgc}{
Institut de Biomécanique Humaine Georges Charpak (IBHGC),
Arts et Métiers ParisTech,
Paris, France
}

\icmlaffiliation{fff}{
French Football Federation (FFF),
Clairefontaine-en-Yvelines, France
}

\icmlcorrespondingauthor{Simon Khan}{simonkhan160@gmail.com}

\icmlkeywords{
3D Human Pose Estimation,
World-Grounded Human Pose,
Sports Computer Vision,
Soccer Broadcast,
Temporal Modeling,
Camera Geometry
}

\vskip 0.3in
]

\printAffiliationsAndNotice{}


\begin{abstract}
Recovering 3D human pose from monocular sports broadcasts remains challenging when players must be localized in a shared metric world coordinate system rather than only reconstructed relative to their own body. We introduce Field Converter, a geometry-initialized temporal residual framework for world-grounded 3D player pose estimation from calibrated soccer broadcasts. Our method first uses camera and pitch geometry to initialize the player root through ray--ground intersection, then predicts a temporal residual correction from pose, image, camera, and geometric cues. On match-disjoint evaluation sequences, residual refinement reduces root error from 49~cm with geometry alone to 14~cm with a frame-wise MLP and 10~cm with a TCN, while a Transformer achieves a comparable 11~cm. The resulting world-space MPJPE reaches 13.2~cm, and ablations show that residual prediction clearly outperforms direct global-root regression while temporal context matters more than the specific temporal backbone. Failure analysis further identifies airborne motion as the main limitation of the ground-based geometric initialization. The code is available at \href{https://github.com/KhanSimon/field_converter}{https://github.com/KhanSimon/field\_converter}.\end{abstract}


\section{Introduction}
\label{sec:introduction}


Recovering the 3D motion of athletes from standard broadcast videos could make large sports video archives much more useful for analysis. Compared with 2D measurements or player-centered 3D pose estimates, reconstructing players in a common metric field coordinate system makes it possible to study how all players move together, measure their relative positions, and relate their motion to the geometry of the pitch. It also provides a basis for combining player motion with ball trajectories and game context. Such representations are relevant to tactical analysis \cite{memmert2019tactical}, augmented coaching \cite{wen2024augmented}
, biomechanical studies \cite{dossantos2021biomechanical}
, injury-risk assessment, officiating \cite{wang2025lasttouch}, and immersive or free-viewpoint replay \cite{hilton2010freeviewpoint}.

Recent advances in monocular 3D human pose estimation and human mesh recovery have substantially improved the estimation of articulated body pose from individual images and videos. Parametric body models such as SMPL~\cite{loper2015smpl} established a compact representation of human pose and shape, while methods such as HybrIK~\cite{li2021hybrik}, 4DHumans~\cite{goel2023humans4d}, Multi-HMR~\cite{baradel2024multihmr}, and SAM 3D Body~\cite{yang2026sam3dbody} have progressively improved robustness to challenging viewpoints, uncommon poses, and multi-person scenes. Temporal approaches such as VideoPose3D~\cite{pavllo2019videopose3d}, MixSTE~\cite{zhang2022mixste}, and MotionBERT~\cite{zhu2023motionbert} further exploit motion context to reduce monocular ambiguities and improve temporal consistency.

However, an accurate body-relative skeleton does not imply accurate localization in the surrounding scene. This distinction is particularly important in team sports. Two players may exhibit very similar local body configurations while occupying completely different regions of the pitch. For tactical or biomechanical applications, the desired representation is therefore not only the articulation of each player relative to the pelvis, but the complete 3D body expressed in a shared and metrically meaningful world coordinate frame.

\begin{figure*}[t]
    \centering
    \includegraphics[width=\textwidth]{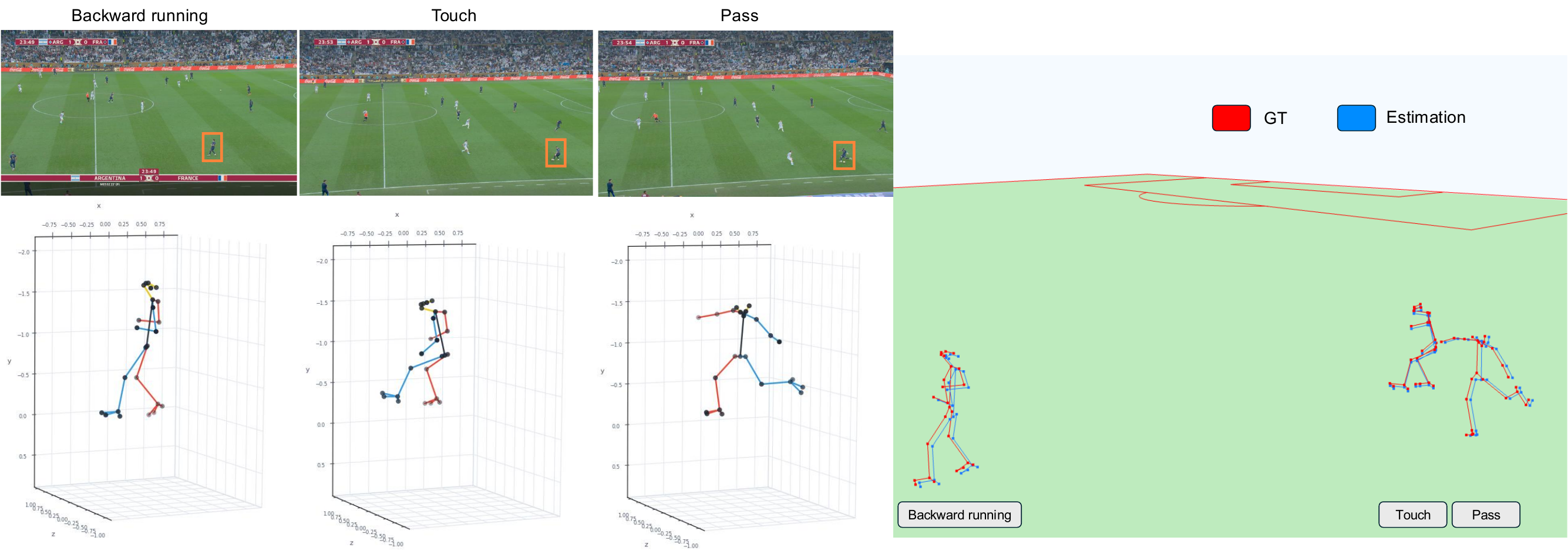}
    \caption{
    From monocular broadcast video to world-grounded 3D player pose. 
    Left: three actions observed in the broadcast and their corresponding camera-relative, self-centered 3D poses. 
    Right: the same reconstructed poses localized in a common metric field coordinate system, where their positions can be directly related to the pitch and to one another.
    }
    \label{fig:teaser}
\end{figure*}

Several recent methods explicitly address metric 3D localization beyond body-relative pose. Ray3D~\cite{zhan2022ray3d} incorporates calibrated camera rays to estimate both 3D body configuration and its metric position with respect to the camera. GLAMR~\cite{yuan2022glamr}, SLAHMR~\cite{ye2023slahmr}, and PACE~\cite{kocabas2024pace} recover human trajectories while accounting for camera motion. WHAM~\cite{shin2024wham}, TRAM~\cite{wang2024tram}, GVHMR~\cite{shen2024gvhmr}, and ProxyCap~\cite{zhang2024proxycap} further introduce temporal, geometric, camera-motion, gravity, and contact priors for reconstruction in a global coordinate system.

The challenge becomes even more pronounced in broadcast soccer. Players are frequently small in the image, undergo partial occlusion, move rapidly over a large metric area, and are observed through a camera undergoing pan, tilt, and zoom. Generic methods for recovering camera motion by tracking static visual features across frames, such as visual odometry and simultaneous localization and mapping (SLAM)~\cite{murartal2015orbslam}, are poorly matched to this setting: the pitch contains large weakly textured regions, players and spectators create substantial dynamic content, and focal length can vary during a broadcast sequence. WorldPose~\cite{jiang2024worldpose}, a large-scale global 3D pose dataset captured during the FIFA World Cup, directly highlights this local-to-global gap and reports large scene-level localization errors for several state-of-the-art global reconstruction approaches.

At the same time, soccer broadcasts contain an unusually strong geometric prior: the pitch has known metric dimensions and standardized line markings. This structure has motivated sports-specific camera-calibration and field-registration methods such as TVCalib~\cite{theiner2023tvcalib}, No Bells, Just Whistles~\cite{gutierrez2024nobells}, PnLCalib~\cite{gutierrez2026pnlcalib}, and BroadTrack~\cite{magera2025broadtrack}. Related work has also shown that known sports geometry can support monocular athlete pose recovery when ordinary scene calibration is difficult~\cite{baumgartner2023partialsports}.

With Field Converter, we exploit the complementary strengths of modern monocular pose estimation and explicit sports geometry. Our starting observation is empirical but important: for the soccer sequences considered here, the relative 3D pose produced by a strong monocular body model is already close to the corresponding camera-relative ground truth after self-centering. The dominant remaining error is therefore not the articulated pose itself, but the global camera-space translation of the player. This motivates a decomposition in which local pose is estimated upstream and our method focuses specifically on world grounding.

This leads to the central question addressed in this work: given monocular broadcast video and calibrated field geometry, how should global player translation be recovered reliably? In particular, is it more effective to refine a geometry-based initialization temporally than to regress global translation directly? To address this question, we first estimate a geometrically grounded root position using ray--ground intersection. We then refine this initialization with a temporal residual network conditioned on pose, image, camera, and field cues. Finally, the refined root anchors the relative skeleton in a common world coordinate system.

%


\section{Related Work}
\label{sec:related_work}


\subsection{Camera-Relative 3D Human Pose and Mesh Recovery}

Modern monocular 3D human reconstruction has progressed from direct joint estimation toward articulated parametric body recovery. SMPL~\cite{loper2015smpl} introduced a widely adopted low-dimensional skinned body model that represents body shape and articulated pose. HybrIK~\cite{li2021hybrik} combines accurate 3D joint prediction with analytical inverse kinematics to obtain body rotations, segment orientations, and human meshes. 4DHumans~\cite{goel2023humans4d} improves image-based reconstruction through a transformer-based HMR architecture and extends the system to tracking in video. Multi-HMR~\cite{baradel2024multihmr} performs multi-person whole-body mesh recovery in a single shot, including 3D localization cues. More recently, SAM 3D Body~\cite{yang2026sam3dbody} introduced a promptable single-image full-body human mesh recovery model designed for robust reconstruction across a wide range of images and poses.

Temporal information is particularly effective for monocular 3D pose estimation because motion provides constraints that are absent in a single image. VideoPose3D~\cite{pavllo2019videopose3d} demonstrated that dilated temporal convolutions can effectively lift 2D keypoint sequences into 3D. Transformer-based approaches such as MixSTE~\cite{zhang2022mixste} model spatial and temporal relations over complete pose sequences, while MotionBERT~\cite{zhu2023motionbert} learns transferable motion representations through large-scale motion pretraining. On the standard Human3.6M benchmark, MotionAGFormer-B~\cite{mehraban2024motionagformer}, for example, reports a Protocol-1 error of $38.4$~mm, illustrating the accuracy achievable by recent temporal 3D pose lifting methods under conventional benchmark conditions.

These approaches mainly target body articulation, camera-relative pose, or camera-centered mesh recovery. Even when temporally smooth and locally accurate, such predictions can still follow an incorrect metric trajectory in the surrounding scene. Our method treats these local estimates as strong upstream observations and focuses specifically on the missing global translation.

\subsection{Absolute and World-Grounded Human Motion Recovery}

A growing body of work seeks to move beyond root-relative reconstruction. Ray3D~\cite{zhan2022ray3d} transforms 2D observations into normalized 3D rays and conditions the prediction on calibrated camera extrinsics, demonstrating the importance of explicit camera geometry for absolute 3D localization.

For dynamic-camera video, GLAMR~\cite{yuan2022glamr} reconstructs global human meshes while accounting for long-term occlusions and camera motion. SLAHMR~\cite{ye2023slahmr} jointly optimizes camera and human trajectories in a shared scene, while PACE~\cite{kocabas2024pace} couples human and camera estimation in a global reconstruction framework. These methods address the ambiguity between observed image motion, human motion, and camera motion.

More recent approaches introduce stronger priors and more efficient formulations. WHAM~\cite{shin2024wham} combines visual and motion features with camera angular velocity and contact-aware trajectory refinement. TRAM~\cite{wang2024tram} estimates camera motion and metric scale before reconstructing a global human trajectory. GVHMR~\cite{shen2024gvhmr} introduces gravity-view coordinates to reduce world-coordinate ambiguity and stabilize world-grounded reconstruction. ProxyCap~\cite{zhang2024proxycap} learns world-space motion from human-centric proxy representations and explicitly targets plausible ground contact.

The difficulty of global localization is particularly apparent on WorldPose~\cite{jiang2024worldpose}. Under its global evaluation, in which a single shared Procrustes transformation aligns the predicted player trajectories to the ground truth, GLAMR and SLAHMR obtain G-MPJPE values of $18{,}888.9$ and $8{,}334.1$~mm, with per-meter drifts of $53.3$ and $17.6$~cm/m, respectively. The same benchmark reports per-player, per-frame Procrustes-aligned PA-MPJPE values of $85.2$ and $163.9$~mm for the two methods. Although these metrics use different alignment protocols and are therefore not directly comparable, they highlight that accurate frame-wise body alignment does not by itself ensure an accurate trajectory in a common scene coordinate system.

Our setting differs from generic in-the-wild reconstruction in an important way: the metric geometry of the soccer pitch is known and the broadcast camera can be calibrated from field markings. We therefore avoid asking the temporal network to infer global translation without structure. Instead, known scene geometry supplies a strong metric initialization, and learning is reserved for the residual error.

\subsection{Sports-Specific 3D Pose Estimation}

Sports video is a challenging domain for monocular pose estimation because athletic motion includes unusual configurations, high accelerations, motion blur, self-occlusion, and subjects that can occupy relatively few image pixels. Sports-specific work has consequently explored domain priors, biomechanical validation, and scene geometry.

Baumgartner and Klatt~\cite{baumgartner2023partialsports} combine 2D pose estimation with partial sports-field registration and jointly reason about athlete pose and camera calibration. Their work shows that line markings and known sports geometry can provide valuable constraints when complete camera calibration is otherwise unavailable.

AutoSoccerPose~\cite{yeung2024autosoccerpose} proposes a semi-automated pipeline for extracting 2D and 3D posture sequences from soccer shooting videos. AthletePose3D~\cite{yeung2025athletepose3d} further quantifies the domain shift associated with athletic motion: on its validation set, using ground-truth 2D poses as input and hip-centered MPJPE, TCPFormer trained only on Human3.6M obtains $234.2$~mm, whereas training on Human3.6M together with AthletePose3D reduces the error to $98.3$~mm. O\v{s}trek et al.~\cite{ostrek2019deployment} compare monocular vision-based motion capture with reference measurements in alpine skiing, highlighting the importance of validation in physically meaningful units rather than relying only on visually plausible motion.

WorldPose~\cite{jiang2024worldpose} is the most directly relevant benchmark to our setting. It provides global player trajectories and body poses from FIFA World Cup footage, together with broadcast-camera information, and is explicitly designed to study multi-person global pose estimation in soccer. In contrast to sports datasets centered primarily on individual athletic motions, WorldPose captures multiple interacting players moving over an entire pitch and therefore directly exposes the local-to-global localization problem addressed in this work.

\subsection{Sports-Field Registration and Broadcast-Camera Estimation}

Metric world reconstruction requires an accurate camera model. In soccer, generic feature-based calibration is difficult because the visible field contains limited texture and the broadcast camera undergoes significant pan, tilt, and zoom. Field markings, however, provide a standardized geometric calibration object.

TVCalib~\cite{theiner2023tvcalib} formulates sports-field registration as camera calibration and optimizes camera parameters using field-segment reprojection. No Bells, Just Whistles~\cite{gutierrez2024nobells} exploits geometric properties of the sports field to improve calibration robustness. PnLCalib~\cite{gutierrez2026pnlcalib} further refines sports-field registration through joint point-and-line optimization. BroadTrack~\cite{magera2025broadtrack} extends the problem temporally and introduces a broadcast-camera tracking system tailored to soccer.

These approaches are complementary to our method. We do not treat camera calibration as the primary contribution. Instead, we assume that camera intrinsics and extrinsics are available from annotations or from an upstream field-registration system, and we use them to construct metric player rays, ground intersections, camera descriptors, and camera-to-world transformations.
%
%
% ...


\section{Materials and Method}

\subsection{Dataset}

We evaluate our method on the FIFA Skeletal Tracking Light 2026 dataset, which contains calibrated monocular soccer-broadcast sequences. Our processed subset comprises 89 clips from eight different matches, totaling approximately 2.41 million valid player--frame observations. The annotations provide time-varying camera intrinsics, extrinsics, and radial-distortion parameters, as well as player bounding boxes, validity masks, and 25-joint 3D skeletons expressed in a common metric field coordinate system.

For each valid player bounding box, we apply SAM 3D Body to obtain 2D keypoints and a 3D body estimate. The resulting skeletons are mapped to the common 25-joint convention, and the sign of the $z$ coordinate is inverted to match the camera-coordinate convention of the dataset. The pelvis is defined as the midpoint of the left and right hip joints, and the resulting 3D skeletons are centered around this point. The camera-space root targets are defined using the same hip midpoint, while the corresponding root-relative ground-truth poses are derived from the annotated world-space skeletons and calibrated camera transformations. The geometry-based root initialization described in Eq.~\eqref{eq:root_init} is computed offline for every valid player--frame observation.

To assess generalization to unseen games, we adopt a strict match-disjoint evaluation protocol. The training set contains 62 clips from six matches, the validation set contains 12 clips from the \texttt{BRA\_KOR} match, and the test set contains 15 clips from the held-out \texttt{ENG\_FRA} match. Consequently, no match appears in more than one split, preventing the model from exploiting match-specific camera trajectories, stadium geometry, or broadcast patterns.

\subsection{Overview of the pipeline}

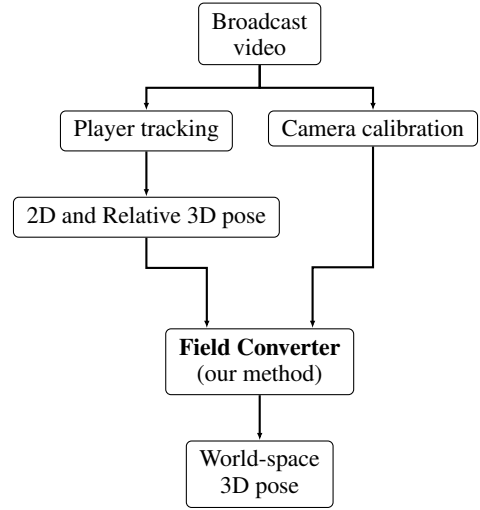
\begin{figure}[t]
    \centering
    \begin{tikzpicture}[
    node distance=6mm and 8mm,
    box/.style={
        draw,
        rounded corners=2pt,
        align=center,
        inner xsep=5pt,
        inner ysep=3.5pt,
        font=\small
    },
    arrow/.style={-{Latex[length=1mm]}, thick}
]

\node[box] (video) {Broadcast\\video};

\node[box, below=of video, xshift=-15mm] (track)
    {Player tracking};

\node[box, below=of track] (pose)
    {2D and Relative 3D pose};

\node[box, below=of video, xshift=15mm] (cam)
    {Camera calibration};

\node[box, below=12mm of pose, xshift=15mm] (world)
    {\textbf{Field Converter}\\(our method)};

\node[box, below=of world] (globalpose)
    {World-space\\3D pose};

\draw[arrow] (video.south) -- ++(0,-3mm) -| (track.north);
\draw[arrow] (video.south) -- ++(0,-3mm) -| (cam.north);

\draw[arrow] (track.south) -- (pose.north);

\draw[arrow]
    (pose.south)
    -- ++(0,-4mm)
    -| ([xshift=-7mm]world.north);

\draw[arrow]
    (cam.south)
    -- ++(0,-16mm)
    -| ([xshift=+7mm]world.north);

\draw[arrow] (world.south) -- (globalpose.north);

\end{tikzpicture}
    \caption{
    Conceptual pipeline. Player tracks, camera calibration, 2D and relative
    3D poses are obtained from the broadcast video and provided to our
    field converter.
    }
    \label{fig:conceptual_pipeline}
\end{figure}

Given a monocular soccer-broadcast sequence, our objective is to recover
the 3D pose of every tracked player in a common metric field coordinate
system. The complete conceptual pipeline is illustrated in
Fig.~\ref{fig:conceptual_pipeline}.

\begin{figure*}[t]
    \centering
    \includegraphics[width=\textwidth]{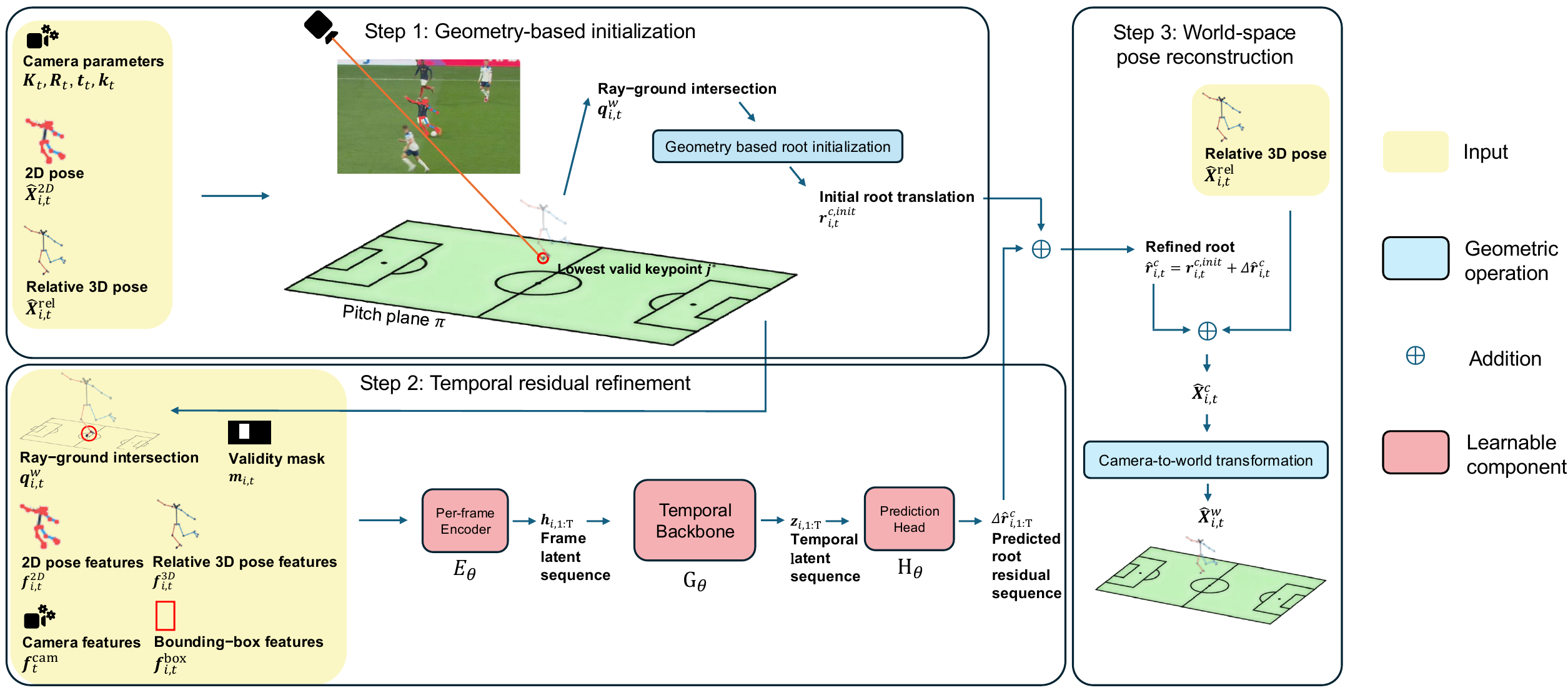}
    \caption{
    Overview of the proposed Field Converter method.
    First, a geometry-based initialization estimates the player root
    from the calibrated camera, the relative 3D pose, and a ray--ground
    intersection. Second, a temporal model predicts a residual correction
    using pose, image, camera, and geometric cues. Finally, the refined
    root anchors the relative skeleton in camera coordinates before
    transformation to the common world coordinate system.
    }
    \label{fig:method_pipeline}
\end{figure*}

The proposed method corresponds to the final world-grounding stage.
For each visible player and frame, we assume access to an upstream
bounding box, 2D keypoints, a self-centered camera-oriented 3D pose,
and calibrated camera parameters. The soccer pitch geometry is known.
Figure~\ref{fig:method_pipeline} details the proposed field-converter
method.

Our method has three stages. First, a lower-limb observation is
back-projected through the calibrated camera and intersected with the
field plane, yielding a geometry-based root initialization. Second, a
temporal network predicts a residual correction to this initialization.
Third, the refined camera-space root anchors the relative skeleton,
which is transformed to the common world coordinate system using the
calibrated camera extrinsics.

\subsection{Notation and Problem Formulation}

Throughout the paper, the first superscript indicates the coordinate representation (e.g., $c$ for camera, $w$ for world, and $\mathrm{rel}$ for pelvis-centered relative coordinates), while a second superscript, when present, indicates the status of the quantity (e.g., $\mathrm{gt}$ for ground truth and $\mathrm{init}$ for geometry-based initialization). Subscripts $i$, $t$, and $j$ refer to the player, frame, and skeletal joint indices, respectively. We use column-vector notation for individual 3D points. A lowercase $\mathbf x_{i,t,j}\in\mathbb R^3$ denotes one skeletal joint, while $\mathbf X_{i,t}\in\mathbb R^{J\times3}$ denotes the corresponding collection of $J$ joints.

The camera at frame $t$ is represented by the intrinsic matrix $\mathbf K_t \in \mathbb R^{3 \times 3}$, a world-to-camera rotation $\mathbf R_t \in SO(3)$, a translation $\mathbf t_t \in \mathbb R^3$, and radial-distortion parameters $\mathbf k_t \in \mathbb R^2$.

A world-space joint $\mathbf x^w_{i,t,j}$ is mapped to camera coordinates by
\begin{equation}
    \mathbf x^c_{i,t,j}
    =
    \mathbf R_t \mathbf x^w_{i,t,j}
    +
    \mathbf t_t,
    \label{eq:world_to_cam}
\end{equation}
and the inverse transformation is
\begin{equation}
    \mathbf x^w_{i,t,j}
    =
    \mathbf R_t^\top
    \left(
        \mathbf x^c_{i,t,j}
        -
        \mathbf t_t
    \right).
    \label{eq:cam_to_world}
\end{equation}
These transformations are applied independently to all $J$ joints.

An upstream monocular body estimator provides a self-centered relative pose
\begin{equation}
    \widehat{\mathbf X}^{\mathrm{rel}}_{i,t}
    \in
    \mathbb R^{J\times3},
\end{equation}
expressed in the camera coordinate frame. In our implementation, the relative skeleton is derived from SAM 3D Body~\cite{yang2026sam3dbody}, mapped to the target joint convention.

The corresponding global camera-space position of each joint is obtained by adding the same root translation to every relative joint:
\begin{equation}
    \widehat{\mathbf x}^{c}_{i,t,j}
    =
    \widehat{\mathbf x}^{\mathrm{rel}}_{i,t,j}
    +
    \widehat{\mathbf r}^{c}_{i,t},
    \qquad j=1,\ldots,J,
    \label{eq:global_cam_pose}
\end{equation}
where $\widehat{\mathbf r}^{c}_{i,t}\in\mathbb R^3$ is the camera-space root translation. Consequently, our learning problem is to recover the global root translation rather than to regenerate the complete articulated body.

For supervised training, each ground-truth world-space joint is transformed to camera coordinates using Eq.~\eqref{eq:world_to_cam}. Let $j_L$ and $j_R$ denote the left- and right-hip joints, respectively. The ground-truth root is defined as their midpoint:
\begin{equation}
    \mathbf r^{c,\mathrm{gt}}_{i,t}
    =
    \frac{1}{2}
    \left(
        \mathbf x^{c,\mathrm{gt}}_{i,t,j_L}
        +
        \mathbf x^{c,\mathrm{gt}}_{i,t,j_R}
    \right).
    \label{eq:gt_root}
\end{equation}
The corresponding ground-truth relative pose is defined joint-wise as
\begin{equation}
    \mathbf x^{\mathrm{rel,gt}}_{i,t,j}
    =
    \mathbf x^{c,\mathrm{gt}}_{i,t,j}
    -
    \mathbf r^{c,\mathrm{gt}}_{i,t},
    \qquad j=1,\ldots,J.
\end{equation}

\subsection{Geometry-Based Root Initialization}
\label{sec:geometry_init}

The playing surface is modeled as a plane
\begin{equation}
    \pi:
    \mathbf n^\top\mathbf x+b=0,
    \label{eq:ground_plane}
\end{equation}
where $\mathbf n$ is the unit field normal vector and $b$ is the plane offset. In the standard soccer-field coordinate system, this can be chosen as $z=0$.

For player $i$ at frame $t$, we select a candidate lower-limb keypoint
\begin{equation}
    j^*_{i,t}
    =
    \arg\max_{j\in\mathcal J_{\mathrm{cand}}}
    v_{i,t,j},
    \label{eq:lowest_joint}
\end{equation}
where image coordinate $v$ increases downward and $\mathcal J_{\mathrm{cand}}$ contains valid candidate joints. In practice, the candidate set can be restricted to ankles, heels, toes, or other lower-limb landmarks.

Let the selected 2D point be
\begin{equation}
    \widetilde{\mathbf p}_{i,t}
    =
    \begin{bmatrix}
    u_{i,t,j^*}\\
    v_{i,t,j^*}\\
    1
    \end{bmatrix}.
\end{equation}
After inverting radial distortion when required, its camera-space ray direction is
\begin{equation}
    \mathbf d^c_{i,t}
    =
    \frac{
        \mathbf K_t^{-1}
        \widetilde{\mathbf p}_{i,t}
    }{
        \left\|
        \mathbf K_t^{-1}
        \widetilde{\mathbf p}_{i,t}
        \right\|_2
    }.
    \label{eq:cam_ray}
\end{equation}

The camera center in world coordinates is
\begin{equation}
    \mathbf C_t
    =
    -
    \mathbf R_t^\top
    \mathbf t_t.
    \label{eq:camera_center}
\end{equation}

The camera viewing direction expressed in world coordinates is obtained by
transforming the positive camera $z$-axis into the world frame:
\begin{equation}
    \mathbf d^{\mathrm{forward}}_t
    =
    \mathbf R_t^\top
    \mathbf e_z,
    \qquad
    \mathbf e_z =
    \begin{bmatrix}
        0 & 0 & 1
    \end{bmatrix}^{\top},
    \label{eq:camera_forward}
\end{equation}
and the world-space ray direction is
\begin{equation}
    \mathbf d^w_{i,t}
    =
    \mathbf R_t^\top
    \mathbf d^c_{i,t}.
\end{equation}
Points along the ray are
\begin{equation}
    \mathbf q^w_{i,t}(\lambda)
    =
    \mathbf C_t
    +
    \lambda
    \mathbf d^w_{i,t}.
\end{equation}

Intersecting the ray with the pitch plane gives
\begin{equation}
    \lambda^*_{i,t}
    =
    -
    \frac{
        \mathbf n^\top\mathbf C_t+b
    }{
        \mathbf n^\top\mathbf d^w_{i,t}
    },
    \label{eq:lambda_ground}
\end{equation}
and
\begin{equation}
    \mathbf q^w_{i,t}
    =
    \mathbf C_t
    +
    \lambda^*_{i,t}
    \mathbf d^w_{i,t}.
    \label{eq:ground_intersection_world}
\end{equation}
The intersection is accepted only when the ray is not parallel to the plane, $\lambda^*_{i,t}>0$, and the required observations are valid.

The intersection is transformed back to camera coordinates:
\begin{equation}
    \mathbf q^c_{i,t}
    =
    \mathbf R_t
    \mathbf q^w_{i,t}
    +
    \mathbf t_t.
\end{equation}
Assuming that the selected joint is in contact with the pitch, the initial root translation is
\begin{equation}
    \mathbf r^{c,\mathrm{init}}_{i,t}
    =
    \mathbf q^c_{i,t}
    -
    \widehat{\mathbf x}^{\mathrm{rel}}_{i,t,j^*}.
    \label{eq:root_init}
\end{equation}

This initialization is metric and directly combines 2D localization, camera geometry, field scale, and the estimated 3D body configuration. It is nevertheless only approximate. If the selected foot is airborne, the ray intersects the pitch at a point behind the true joint location. Occlusion, motion blur, imperfect keypoint localization, and relative-pose error can produce similar biases. These failure modes motivate a learned residual correction.

\subsection{Input Representation}

For each player and frame, the temporal model receives a feature vector
\begin{equation}
\mathbf f_{i,t}
=
\left[
\mathbf f^{3D}_{i,t},
\mathbf f^{2D}_{i,t},
\mathbf f^{\mathrm{box}}_{i,t},
\mathbf f^{\mathrm{cam}}_t,
\mathbf q^{\mathrm{w}}_{i,t},
\mathbf m_{i,t}
\right].
\label{eq:features}
\end{equation}

\paragraph{Relative 3D pose.}
The relative-pose descriptor is the flattened pelvis-centered skeleton:
\begin{equation}
    \mathbf f^{3D}_{i,t}
    =
    \operatorname{vec}
    \left(
    \widehat{\mathbf X}^{\mathrm{rel}}_{i,t}
    \right).
\end{equation}

\paragraph{2D pose.}
We include both image-normalized and bounding-box-normalized 2D keypoints. For an image of width $W$ and height $H$,
\begin{equation}
    \mathbf p^{\mathrm{img}}_{i,t,j}
    =
    \left[
    \frac{u_{i,t,j}}{W},
    \frac{v_{i,t,j}}{H}
    \right].
\end{equation}
Given box center $(c^x_{i,t},c^y_{i,t})$, width $w_{i,t}$, and height $h_{i,t}$,
\begin{equation}
    \mathbf p^{\mathrm{box}}_{i,t,j}
    =
    \left[
    \frac{u_{i,t,j}-c^x_{i,t}}{w_{i,t}},
    \frac{v_{i,t,j}-c^y_{i,t}}{h_{i,t}}
    \right].
\end{equation}
We then define the 2D pose descriptor as the concatenation of both representations over all joints:
\begin{equation}
\mathbf f^{2D}_{i,t}
=
\operatorname{vec}
\left(
\left[
\mathbf p^{\mathrm{img}}_{i,t,j},
\mathbf p^{\mathrm{box}}_{i,t,j}
\right]_{j=1}^{J}
\right).
\end{equation}

\paragraph{Bounding-box geometry.}
The normalized box descriptor is
\begin{equation}
\mathbf f^{\mathrm{box}}_{i,t}
=
\left[
\frac{c^x_{i,t}}{W},
\frac{c^y_{i,t}}{H},
\frac{w_{i,t}}{W},
\frac{h_{i,t}}{H},
\log\frac{w_{i,t}}{h_{i,t}}
\right].
\label{eq:bbox_features}
\end{equation}
Although the body estimator itself may already use the bounding box, box scale and image position remain useful cues for metric depth and perspective.

\paragraph{Camera features.}
The camera descriptor contains normalized intrinsic parameters, lens distortion, camera position, and viewing direction:
\begin{equation}
\mathbf f^{\mathrm{cam}}_t
=
\left[
\frac{f_{x,t}}{W},
\frac{f_{y,t}}{H},
\frac{c_{x,t}}{W},
\frac{c_{y,t}}{H},
k_{1,t},
k_{2,t},
\mathbf C_t,
\mathbf d^{\mathrm{forward}}_t
\right].
\label{eq:camera_features}
\end{equation}

\paragraph{Ground-intersection feature.}
The geometric feature provided to the temporal model is the world-space ray--ground intersection
\begin{equation}
    \mathbf q^{\mathrm{w}}_{i,t} \in \mathbb{R}^{3},
\end{equation}
obtained from the selected lower-body keypoint as described in Sec.~\ref{sec:geometry_init}. This feature directly provides the network with the metric ground location implied by the current image observation and camera geometry.

\paragraph{Validity masks.}
Joint validity is encoded for each player and frame by
\begin{equation}
    \mathbf m_{i,t}
    =
    \left[
        m_{i,t,1}, \ldots, m_{i,t,J}
    \right]
    \in \{0,1\}^{J},
\end{equation}
where $m_{i,t,j}=1$ indicates that joint $j$ is valid and
$m_{i,t,j}=0$ otherwise. This joint-level mask is provided explicitly as an input feature.

We separately denote by $v_{i,t}\in\{0,1\}$ the player--frame validity indicator used to mask the training objectives.

\subsection{Temporal Residual Refinement}

We define the ground-truth residual as
\begin{equation}
    \Delta\mathbf r^{c,\mathrm{gt}}_{i,t}
    =
    \mathbf r^{c,\mathrm{gt}}_{i,t}
    -
    \mathbf r^{c,\mathrm{init}}_{i,t}.
    \label{eq:gt_residual}
\end{equation}

The temporal refinement model $F_\theta$ predicts this residual from the
sequence of input features:
\begin{equation}
    \Delta\widehat{\mathbf r}^{c}_{i,1:T}
    =
    F_\theta
    \left(
    \mathbf f_{i,1:T}
    \right),
\end{equation}
where $\Delta\widehat{\mathbf r}^{c}_{i,t}$ denotes the predicted root
correction in camera coordinates.

A per-frame encoder first maps the input vector into a latent representation:
\begin{equation}
    \mathbf h_{i,t}
    =
    E_\theta
    \left(
    \mathbf f_{i,t}
    \right).
\end{equation}
A temporal backbone then aggregates information across the sequence:
\begin{equation}
    \mathbf z_{i,1:T}
    =
    G_\theta
    \left(
    \mathbf h_{i,1:T}
    \right).
\end{equation}
Finally, a lightweight prediction head estimates the residual:
\begin{equation}
    \Delta\widehat{\mathbf r}^{c}_{i,t}
    =
    H_\theta
    \left(
    \mathbf z_{i,t}
    \right).
\end{equation}

Thus, $F_\theta$ denotes the complete refinement model,
i.e. the composition of $E_\theta$, $G_\theta$, and $H_\theta$. The refined root is
\begin{equation}
    \widehat{\mathbf r}^{c}_{i,t}
    =
    \mathbf r^{c,\mathrm{init}}_{i,t}
    +
    \Delta\widehat{\mathbf r}^{c}_{i,t}.
    \label{eq:refined_root}
\end{equation}

 The temporal backbone can be instantiated with a temporal convolutional network or a Transformer. A TCN uses stacked one-dimensional convolutions, residual connections, and dilation to obtain a large temporal receptive field with low computational cost. A Transformer instead uses positional information and bidirectional self-attention to model long-range dependencies. The residual formulation is independent of this architectural choice. Table~\ref{tab:refinement_architecture} describes the architecture used.
\begin{table}[t]
\centering
\caption{Architecture of the temporal residual refinement models.}
\label{tab:refinement_architecture}
\resizebox{\columnwidth}{!}{%
    \begin{tabular}{lcc}
\toprule
Component & TCN & Transformer \\
\midrule
Input dimension       & 370 & 345 \\
Frame encoder         & 370--192--192--192 & 345--256--256 \\
Latent dimension      & 192 & 256 \\
Temporal layers       & 5 residual blocks & 2 encoder layers \\
Temporal operator     & $2\times$ Conv1D/block & Self-attention \\
Kernel / heads        & $k=3$ & 4 heads \\
Dilations / FFN width & $(1,2,4,8,16)$ & 512 \\
Normalization         & None & Pre-LayerNorm \\
Positional encoding   & Not required & Learned \\
Prediction head       & 192--128--3 & 256--128--3 \\
Activation            & GELU & GELU \\
Dropout               & 0.14 & 0.10 \\
Window / stride       & 41 / 8 frames & 41 / 8 frames \\
Parameters            & 1.278M & 1.252M \\
\bottomrule
\end{tabular}
}
\end{table}

Training and inference use overlapping temporal windows. If a frame belongs to several windows, the corresponding root estimates are aggregated:
\begin{equation}
    \widehat{\mathbf r}^{c}_{i,t}
    =
    \frac{1}{
        \left|\mathcal W_{i,t}\right|
    }
    \sum_{w\in\mathcal W_{i,t}}
    \widehat{\mathbf r}^{c,(w)}_{i,t},
    \label{eq:window_aggregation}
\end{equation}
where $\mathcal W_{i,t}$ denotes the set of windows containing frame $t$.

\subsection{Training Objectives}

The primary objective supervises the refined camera-space root. Let
$N_r=\sum_{i,t}v_{i,t}$ denote the number of valid player--frame
observations and $\Omega=\sum_{a\in\{x,y,z\}}\omega_a$ the sum of the
axis weights. We use the axis-weighted Smooth L1 loss
\begin{equation}
\mathcal L_{\mathrm{root}}
=
\frac{1}{N_r\Omega}
\sum_{i,t} v_{i,t}
\sum_{a\in\{x,y,z\}}
\omega_a\,
\rho_{\beta}\!\left(
\widehat r^{c}_{i,t,a}
-
r^{c,\mathrm{gt}}_{i,t,a}
\right).
\label{eq:root_loss}
\end{equation}
Here, $v_{i,t}$ is the player--frame validity mask and $\omega_a$
controls the contribution of axis $a$. The scalar Smooth L1 penalty is
\begin{equation}
\rho_{\beta}(e)=
\begin{cases}
\dfrac{e^2}{2\beta}, & |e| < \beta,\\[4pt]
|e|-\dfrac{\beta}{2}, & |e|\geq\beta,
\end{cases}
\qquad \beta=1.
\label{eq:smooth_l1}
\end{equation}
The penalty is applied independently to each coordinate. Root,
velocity, and acceleration losses are evaluated in normalized root
coordinates. The camera-space consistency loss is evaluated in metres.

To supervise temporal dynamics, we define the predicted and
ground-truth first-order root differences as
\begin{equation}
\widehat{\mathbf v}_{i,t}
=
\widehat{\mathbf r}^{c}_{i,t}
-
\widehat{\mathbf r}^{c}_{i,t-1},
\end{equation}
and
\begin{equation}
\mathbf v^{\mathrm{gt}}_{i,t}
=
\mathbf r^{c,\mathrm{gt}}_{i,t}
-
\mathbf r^{c,\mathrm{gt}}_{i,t-1}.
\end{equation}
The first order temporal loss is then
\begin{equation}
\mathcal L_{\mathrm{vel}}
=
\frac{1}{3N_v}
\sum_{i,t}
m^{v}_{i,t}
\sum_{a\in\{x,y,z\}}
\rho_{\beta}\!\left(
\widehat v_{i,t,a}
-
v^{\mathrm{gt}}_{i,t,a}
\right),
\end{equation}
where $m^{v}_{i,t}$ is one only when both consecutive frames are valid,
and $N_v=\sum_{i,t}m^{v}_{i,t}$ is the corresponding number of valid
frame pairs.

A second-order temporal difference is defined as
\begin{equation}
\widehat{\mathbf a}_{i,t}
=
\widehat{\mathbf r}^{c}_{i,t}
-
2\widehat{\mathbf r}^{c}_{i,t-1}
+
\widehat{\mathbf r}^{c}_{i,t-2},
\end{equation}
with the corresponding ground-truth quantity
$\mathbf a^{\mathrm{gt}}_{i,t}$. The second order temporal loss is
\begin{equation}
\mathcal L_{\mathrm{acc}}
=
\frac{1}{3N_a}
\sum_{i,t}
m^{a}_{i,t}
\sum_{a\in\{x,y,z\}}
\rho_{\beta}\!\left(
\widehat a_{i,t,a}
-
a^{\mathrm{gt}}_{i,t,a}
\right),
\end{equation}
where $m^{a}_{i,t}$ requires the three consecutive frames to be valid,
and $N_a=\sum_{i,t}m^{a}_{i,t}$.

The reconstructed camera-space pose is
\begin{equation}
\widehat{\mathbf X}^{c}_{i,t,j}
=
\widehat{\mathbf X}^{\mathrm{rel}}_{i,t,j}
+
\widehat{\mathbf r}^{c}_{i,t}.
\end{equation}
A masked 3D consistency loss compares the reconstructed joints with
the full camera-space ground truth. Let
\[
N_j=\sum_{i,t,j} v_{i,t}m_{i,t,j}
\]
denote the number of valid player--frame--joint observations. We use
\begin{equation}
\begin{aligned}
\mathcal L_{\mathrm{cam3D}}
&=
\frac{1}{3N_j}
\sum_{i,t,j}
v_{i,t}m_{i,t,j} \\
&\quad \times
\sum_{a\in\{x,y,z\}}
\rho_{1}\!\left(
\widehat X^{c}_{i,t,j,a}
-
X^{c,\mathrm{gt}}_{i,t,j,a}
\right).
\end{aligned}
\label{eq:cam3d_loss}
\end{equation}
Here, $m_{i,t,j}$ is the joint-validity mask and $v_{i,t}$ is the
player--frame validity mask. Since the relative skeleton is not
refined, this term provides only auxiliary supervision.

The overall objective is
\begin{equation}
\mathcal L
=
\lambda_{\mathrm{root}}
\mathcal L_{\mathrm{root}}
+
\lambda_{\mathrm{vel}}
\mathcal L_{\mathrm{vel}}
+
\lambda_{\mathrm{acc}}
\mathcal L_{\mathrm{acc}}
+
\lambda_{\mathrm{cam3D}}
\mathcal L_{\mathrm{cam3D}}.
\label{eq:full_loss}
\end{equation}

Both models are optimized with AdamW using an initial learning rate of
$1.7\times10^{-4}$ and a weight decay of
$3.2\times10^{-4}$. No learning-rate scheduler is used. Training
uses batches of 64 temporal windows for at most 60 epochs. Gradients
are clipped to a maximum global $\ell_2$ norm of 1. Model selection is
based on the lowest mean root error on the validation set, with early
stopping after 10 consecutive epochs without improvement.Table~\ref{tab:training_configuration} describes the training configuration that we used. 

\begin{table}[t]
\centering
\caption{Training configuration of the selected temporal refinement models.}
\label{tab:training_configuration}
\resizebox{\columnwidth}{!}{%
    \begin{tabular}{lcc}
\toprule
Training setting & TCN & Transformer \\
\midrule
Optimizer
    & AdamW & AdamW \\
Initial learning rate
    & $1.6788\times10^{-4}$ & $1.6788\times10^{-4}$ \\
Weight decay
    & $3.2001\times10^{-4}$ & $3.2001\times10^{-4}$ \\
Learning-rate scheduler
    & None & None \\
Batch size
    & 64 windows & 64 windows \\
Maximum epochs
    & 60 & 60 \\
Early-stopping patience
    & 10 epochs & 10 epochs \\
Selection criterion
    & Validation root error & Validation root error \\
Gradient clipping
    & $\ell_2$ norm, max. 1.0 & $\ell_2$ norm, max. 1.0 \\
Best epoch
    & 41 & 23 \\
Last training epoch
    & 51 & 33 \\
\bottomrule
\end{tabular}
}
\end{table}

\subsection{Hyperparameter Selection}

Hyperparameter selection was first conducted using the TCN. We performed a 12-trial random search over architectural, temporal,
optimization, and loss-related parameters. Each trial was ranked using
the mean root translation error on the validation set. Starting from
the best configuration, focused grid searches evaluated six
window--stride combinations and 16 combinations of auxiliary loss
weights.

The Transformer was optimized independently using a multi-fidelity
random-search procedure. We evaluated 16 configurations, comprising the
reference configuration and 15 randomly sampled alternatives. The search
varied the temporal window and stride, encoder and latent dimensions,
number of self-attention layers and heads, feed-forward dimension,
dropout rate, positional encoding, prediction-head dimension, learning
rate, and weight decay. The loss weights, batch size, gradient-clipping
threshold, and early-stopping criterion were kept fixed during this
search.

\subsection{World-Space Reconstruction}

Once the root has been refined and denormalized, each global
camera-space joint is obtained using Eq.~\eqref{eq:global_cam_pose}.
The final world-space position of joint $j$ is
\begin{equation}
    \widehat{\mathbf x}^{w}_{i,t,j}
    =
    \mathbf R_t^\top
    \left(
        \widehat{\mathbf x}^{c}_{i,t,j}
        -
        \mathbf t_t
    \right),
    \qquad j=1,\ldots,J.
    \label{eq:world_reconstruction}
\end{equation}
All players are therefore represented in the same metric soccer-field coordinate system.

If an upstream body mesh is available, the same translation can be applied to every relative mesh vertex. For vertex $k$,
\begin{equation}
    \widehat{\mathbf v}^{c}_{i,t,k}
    =
    \widehat{\mathbf v}^{\mathrm{rel}}_{i,t,k}
    +
    \widehat{\mathbf r}^{c}_{i,t},
\end{equation}
followed by
\begin{equation}
    \widehat{\mathbf v}^{w}_{i,t,k}
    =
    \mathbf R_t^\top
    \left(
        \widehat{\mathbf v}^{c}_{i,t,k}
        -
        \mathbf t_t
    \right).
\end{equation}
This provides world-grounded meshes for visualization or free-viewpoint rendering while leaving the original relative body shape and articulation unchanged.

\subsection{Evaluation Protocol}
\label{sec:evaluation_protocol}

We evaluate global localization using the Euclidean error of the predicted camera-space root (Root error) and the mean per-joint position error after reconstruction in the common field coordinate system (World MPJPE). We additionally report Local MPJPE, obtained after removing the global root translation, to separate errors originating from the upstream relative pose estimator from errors due to global localization. Finally, reprojection error measures the mean 2D distance, in pixels, between the reconstructed 3D joints projected through the calibrated camera and their reference 2D locations. All metrics are computed on valid player--frame and joint observations only.

Since all compared root-refinement models use the same upstream relative 3D pose, Local MPJPE is identical by construction across the geometry, MLP, TCN, and Transformer variants. This metric is therefore primarily reported to quantify the remaining local-pose error independently of the global localization stage.


\section{Results}
\label{sec:results}

\subsection{Main Quantitative Results}
\label{sec:main_results}

\paragraph{Results.}
Table~\ref{tab:main_results} summarizes the main quantitative results.
The geometry-based initialization yields a root error of $49$~cm.
Learning a residual correction substantially improves this estimate:
the frame-wise MLP reduces the root error to $14$~cm.
Adding temporal context further improves global localization, with
root errors of $10$~cm for the TCN and $11$~cm for the Transformer.
The corresponding World MPJPE is $13$~cm for both temporal models,
compared with $16$~cm for the frame-wise MLP.
The Local MPJPE remains $8$~cm for all variants.

\paragraph{Interpretation.}
Residual learning reduces the root error by approximately $72\%$ relative
to the geometry-based initialization even without temporal modeling.
The additional improvement obtained by the TCN and Transformer shows
that temporal information is beneficial for global player localization.
However, the nearly identical World MPJPE obtained by the two temporal
architectures suggests that the availability of temporal context is more
important than the specific choice between convolutional and
attention-based modeling.
The unchanged Local MPJPE further confirms that the performance gains
originate from improved global localization rather than changes to the
relative body pose.

\begin{table*}[t]
    \centering
    \caption{
    Main quantitative results. All learning-based models predict a residual
    correction to the geometry-based root initialization. The relative 3D
    pose is shared across all methods and is therefore unchanged by root
    refinement.
    }
    \label{tab:main_results}
    \resizebox{\textwidth}{!}{%
        \begin{tabular}{lrrrrr}
\toprule
Method & Root error $\downarrow$ (cm) & World MPJPE $\downarrow$ (cm) & Local MPJPE $\downarrow$ (cm) & Reproj. $\downarrow$ (px) & Params. (M) \\
\midrule
Geometry initialization & 48.58 & 48.36 & 7.74 & 5.39 & 0\\
MLP residual & 13.63 & 15.76 & 7.74 & 3.69 & 0.194 \\
TCN residual (41 frames) & 10.12 & 13.20 & 7.74 & 3.49 & 1.278 \\
Transformer residual (41 frames) & 11.04 & 13.21 & 7.74 & 3.43 & 1.252\\
\bottomrule
\end{tabular}

    }
\end{table*}

\subsection{Ablation Studies}
\label{sec:ablations}

\subsubsection{Absolute versus Residual Root Prediction}
\label{sec:ablation_residual}

\paragraph{Results.}
Table~\ref{tab:direct_vs_residual} compares direct absolute-root
prediction with the proposed residual formulation.
Direct root prediction results in errors of $2.56$~m for the frame-wise
MLP and $63$~cm for the TCN.
The geometry-only initialization reaches $49$~cm.
When the same learning models instead predict a residual correction
with respect to this initialization, the errors decrease to $14$~cm
for the MLP and $10$~cm for the TCN.

\paragraph{Interpretation.}
The large performance gap between direct and residual prediction
supports the proposed decomposition of the problem.
Camera and field geometry provide a strong metric prior, considerably
reducing the range of translations that must be inferred from visual
observations.
The learning model can therefore focus on estimating the
context-dependent error of this initialization rather than recovering
the complete global player position from scratch.

\begin{table}[t]
    \centering
    \caption{
    Effect of the prediction target. Direct global-root regression is
    compared with geometry-based initialization and residual
    ($\Delta$ root) refinement.
    }
    \label{tab:direct_vs_residual}
    \resizebox{\columnwidth}{!}{%
        \begin{tabular}{lll r}
\toprule
\textbf{Initialization} & \textbf{Target} & \textbf{Temporal} & \textbf{Root error} $\downarrow$ (cm) \\
\midrule
---      & absolute root  & MLP & 256.00 \\
---      & absolute root  & TCN & 63.05 \\
Geometry & ---            & --- & 48.58 \\
Geometry & $\Delta$ root  & MLP & 13.63 \\
Geometry & $\Delta$ root  & TCN & \textbf{10.12} \\
\bottomrule
\end{tabular}
    }
\end{table}

\subsubsection{Input Cues}
\label{sec:ablation_inputs}
\paragraph{Results.}
Table~\ref{tab:input_ablation} reports the contribution of each input
modality for the TCN model.
Removing the valid-joint mask increases root error from $10$ to
$11$~cm.
Removing the ray--ground intersection also results in an error of
approximately $11$~cm.
Without the relative 3D pose, root error increases to $12$~cm, while
removing the camera descriptor increases it to $13$~cm.
The strongest degradation is observed when removing the 2D pose and
bounding-box cues, yielding a root error of $16$~cm and a World MPJPE
of $18$~cm.

\paragraph{Interpretation.}
No single input modality fully determines global player localization.
The explicit ray--ground intersection remains useful even though it is
already involved in constructing the initial root, indicating that the
network benefits from direct access to the underlying geometric cue.
Relative 3D pose and camera information provide complementary
information about body configuration and scene geometry.
The particularly large degradation observed without 2D pose and
bounding-box features shows that image-space position and apparent
player scale remain important cues for correcting metric depth and
global translation.

\begin{table*}[t]
    \centering
    \caption{
    Input ablation for the TCN residual model. $\Delta$ Root denotes the
    increase in root error relative to the full model.
    }
    \label{tab:input_ablation}
    \resizebox{\textwidth}{!}{%
        \begin{tabular}{llrrrr}
\toprule
Model & Input configuration & Root $\downarrow$ (cm) & $\Delta$ Root (cm) & World MPJPE $\downarrow$ (cm) & Reproj. $\downarrow$ (px) \\
\midrule
TCN & Full & 10.12 & 0.00 & 13.20 & 3.49 \\
TCN & w/o valid-joint mask $\mathbf{m}_{i,t}$ & 10.75 & 0.63 & 13.53 & 3.41 \\
TCN & w/o ground intersection $\mathbf{q}^{w}_{i,t}$ & 11.17 & 1.05 & 13.87 & 3.43 \\
TCN & w/o relative 3D pose $\mathbf{f}^{3D}$ & 12.40 & 2.28 & 14.93 & 3.70 \\
TCN & w/o 2D pose and box cues $\mathbf{f}^{2D},\mathbf{f}^{\mathrm{box}}$ & 15.93 & 5.81 & 18.20 & 5.54 \\
TCN & w/o camera descriptor $\mathbf{f}^{cam}$ & 13.02 & 2.90 & 15.39 & 3.50 \\
\bottomrule
\end{tabular}
    }
\end{table*}

\subsection{Analysis of the Geometric Refinement}
\label{sec:analysis}

\subsubsection{Robustness to Initialization Error}
\label{sec:initialization_analysis}
\begin{figure}[t]
    \centering
    \includegraphics[width=\columnwidth]
    {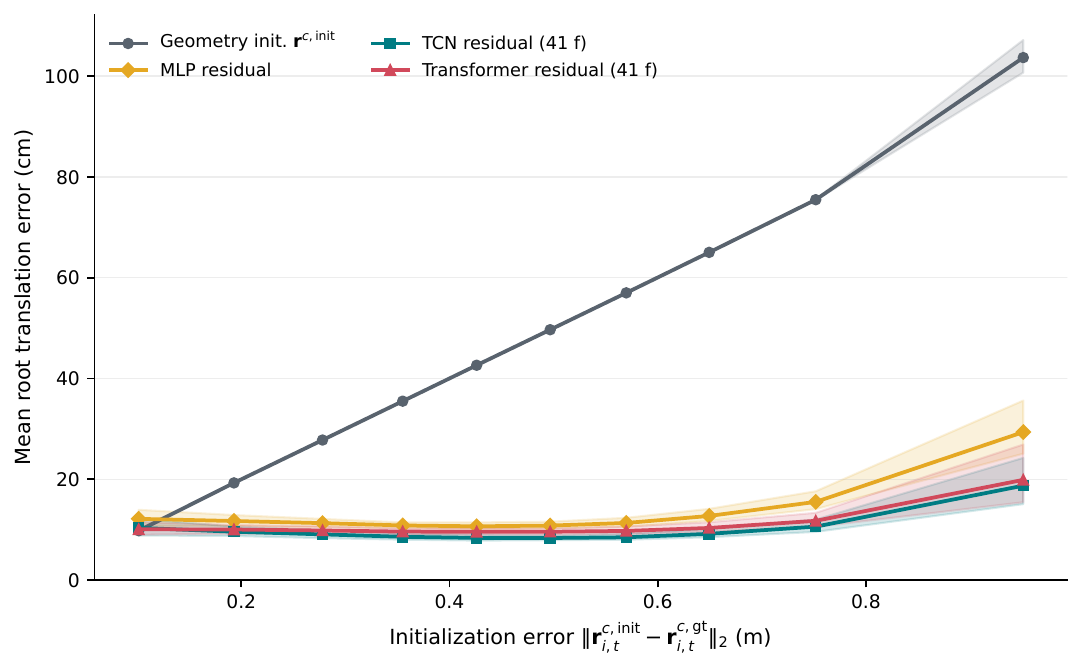}
    \caption{
    Root error as a function of the geometry-based initialization error.
    Temporal residual refinement remains substantially more stable than
    the raw geometric estimate as initialization quality deteriorates.
    }
    \label{fig:error_vs_init}
\end{figure}

\paragraph{Results.}
Figure~\ref{fig:error_vs_init} reports final root error as a function
of the geometry-based initialization error.
For already accurate geometric estimates, the learned correction
provides little improvement and can slightly perturb an initialization
that is already close to the ground truth.
As initialization error increases, however, the temporal models remain
substantially more stable.
The TCN maintains a root error of approximately $8$--$11$~cm over a
broad range of initialization errors, before increasing to approximately
$19$~cm in the most difficult bin, where the geometric initialization
exceeds one meter on average.

\paragraph{Interpretation.}
The residual model is therefore not limited to producing small local
adjustments around the geometric estimate.
When the initialization becomes strongly biased, the network can use
pose, image, camera, and temporal information to recover a substantially
more accurate global position.
Conversely, the small gain obtained when the initialization is already
accurate suggests that most of the benefit of learning is concentrated
on geometrically ambiguous or incorrectly grounded configurations.

\subsubsection{Airborne Players and Failure Cases}
\label{sec:airborne_analysis}

\begin{figure*}[t]
    \centering

    \subfigure[
        Root error as a function of lower-foot clearance above the pitch.
        \label{fig:foot_clearance}
    ]{
        \includegraphics[width=0.57\textwidth]
        {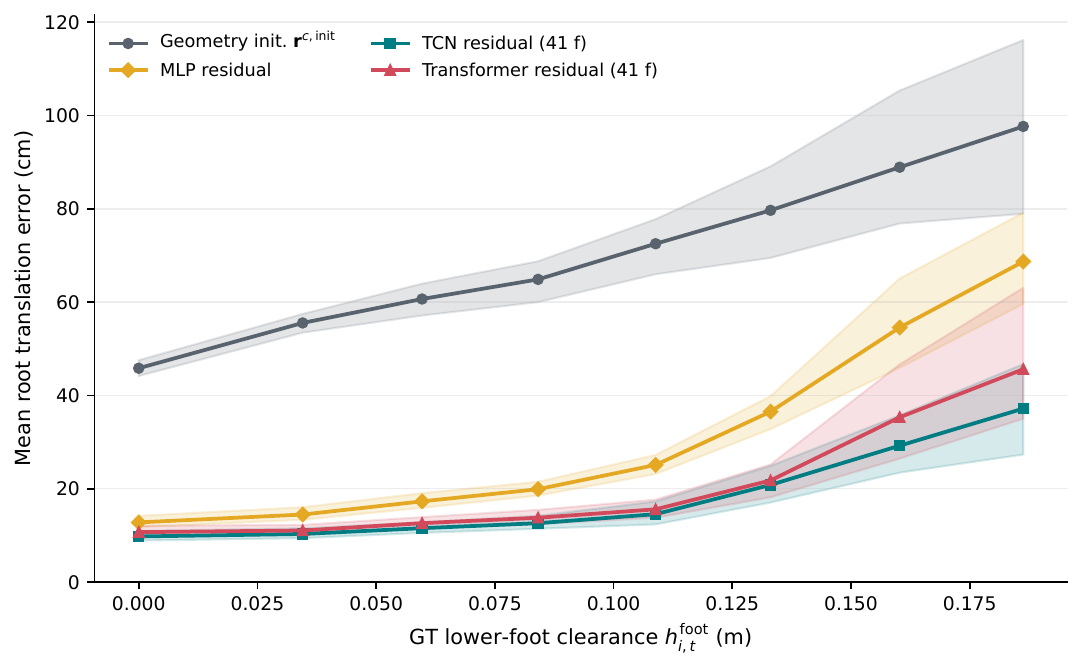}
    }
    \hfill
    \subfigure[
        Qualitative airborne failure case during a header.
        Ground truth is shown in red and the reconstructed pose in orange.
        \label{fig:airborne_qualitative}
    ]{
        \includegraphics[width=0.39\textwidth]
        {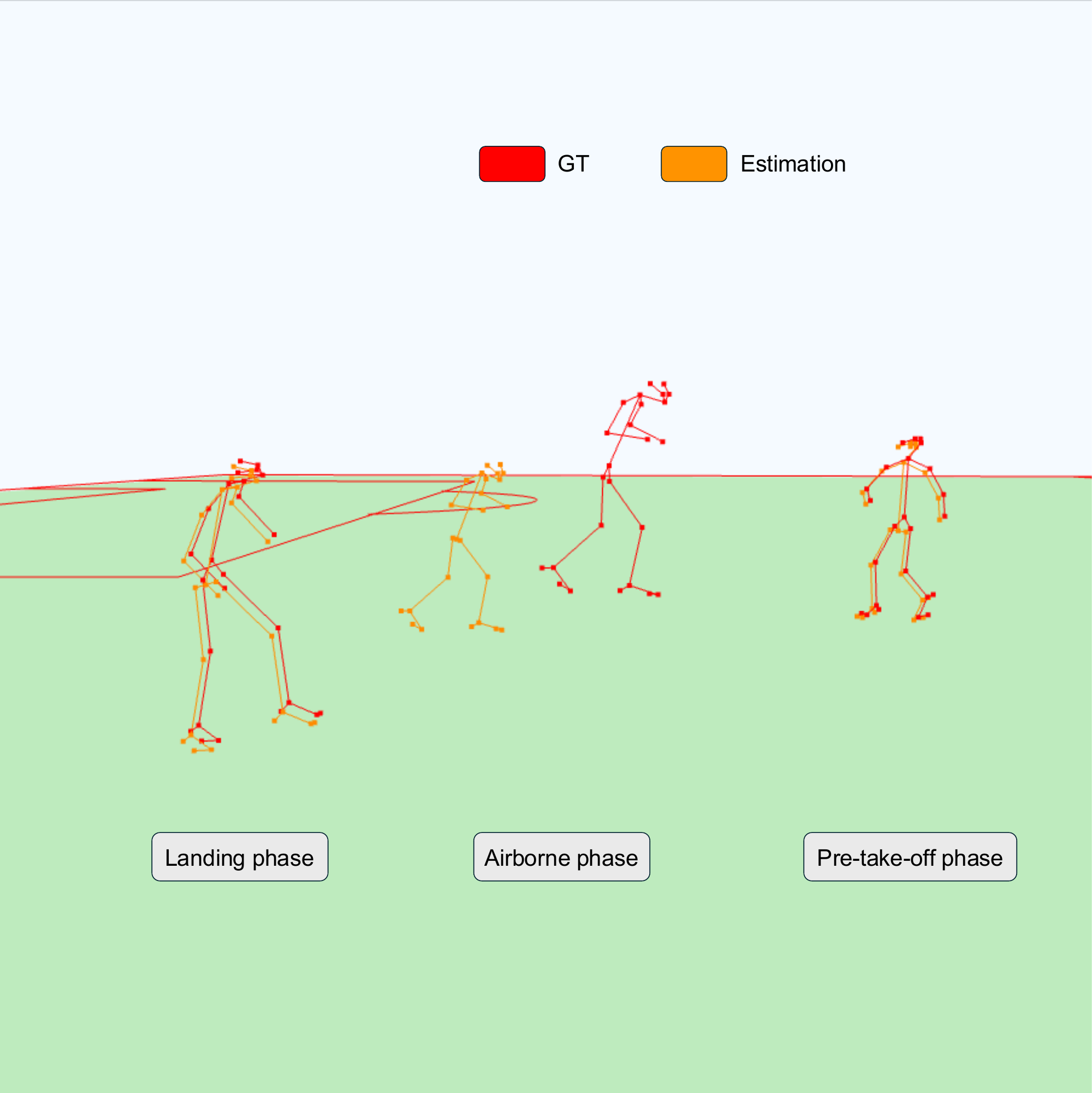}
    }

    \caption{
    Airborne-player analysis.
    The geometric assumption is increasingly violated as both feet move away
    from the playing surface, producing larger initialization and final
    localization errors.
    Temporal refinement mitigates, but does not entirely remove, this failure
    mode.
    }
    \label{fig:airborne_failure}
\end{figure*}

\paragraph{Results.}
Figure~\ref{fig:airborne_failure}a reports root error as a function of
the clearance of the lower foot from the playing surface.
When the lower foot is close to the ground, the TCN achieves
approximately $10$~cm root error.
The error progressively increases with foot clearance, reaching
approximately $15$~cm at $10$--$13$~cm clearance, $29$~cm at
$15$--$18$~cm, and $37$~cm in the highest-clearance interval.
The same trend is considerably stronger for the geometry-based
initialization, whose error increases from approximately $46$~cm for
grounded configurations to almost one meter for the largest foot
clearances.
Figure~\ref{fig:airborne_failure}b illustrates an example during a
heading action, where the predicted pose remains locally plausible but
is globally displaced relative to the ground truth.

\paragraph{Interpretation.}
These results expose a direct limitation of the geometric initialization.
The ray--ground construction implicitly assumes that the selected
lower-limb joint lies close to the playing surface.
When the player becomes airborne, the corresponding image ray intersects
the pitch behind the true 3D joint location, producing a systematically
biased root estimate.
Temporal refinement compensates for a substantial part of this error,
but does not completely remove the underlying ambiguity.
Airborne motion therefore remains one of the main failure modes of the
current approach and motivates future refinements that explicitly model
ground contact or airborne states.

\subsection{Additional Analyses}
\label{sec:additional_analyses}

\paragraph{Results.}
Additional quantitative analyses are provided in the supplementary
material. They include comparisons of model complexity and temporal
consistency, performance across different motion regimes, sensitivity
to image location, player scale and speed, and a decomposition of root
localization error along the camera axes.

\paragraph{Interpretation.}
These complementary analyses further characterize the behavior of the
proposed refinement model without overloading the main paper.
In particular, they show that temporal modeling is especially beneficial
in difficult motion regimes and that the dominant remaining localization
uncertainty is associated with camera depth.


\section{Conclusion}
\label{sec:conclusion}

We introduced Field Converter, a geometry-initialized temporal residual framework for world-grounded 3D player pose estimation from monocular soccer broadcasts. Rather than regressing global player position directly, the proposed approach first exploits calibrated camera and pitch geometry to obtain a metric root initialization and then learns only its temporal residual correction. This decomposition substantially improves global localization while preserving the relative 3D pose estimated upstream. Our experiments show that residual prediction is markedly more effective than direct root regression, that temporal modeling further improves both localization accuracy and trajectory consistency, and that TCN and Transformer backbones achieve comparable performance. The ablation study further confirms that image-space pose, bounding-box, camera, relative-pose, and ray--ground cues provide complementary information. Overall, these results support the central premise of this work: in this broadcast-soccer setting, accurate world-grounded reconstruction depends less on re-estimating local body articulation than on robustly recovering the global translation that anchors this articulation to the field.

The current approach nevertheless relies on several assumptions that leave room for improvement. In particular, the geometry-based initialization assumes that the selected lower-limb joint lies close to the playing surface, which becomes invalid during airborne actions such as jumps and headers and leads to substantially larger localization errors. The method also depends on upstream player detections, pose estimates, and camera calibration, so errors in these components may propagate to the final reconstruction. Future work will therefore investigate explicit modeling of ground-contact and airborne states, improved geometric initialization when no reliable ground contact is available, and training strategies that better represent rare high-clearance motions.


\bibliography{references}
\bibliographystyle{icml2021}



\clearpage
\appendix

\section{Supplementary Material}
\label{sec:supplementary}

This supplementary material provides additional quantitative analyses of the proposed root-refinement framework. We report complementary comparisons of model architecture and temporal consistency, extend the input ablation to both temporal backbones, and analyze performance as a function of motion and observation conditions.

\subsection{Architecture and Model Complexity}
\label{sec:supp_architecture}

Table~\ref{tab:supp_architecture} compares the frame-wise MLP, TCN, and Transformer residual models in terms of localization accuracy and number of trainable parameters.

The MLP contains substantially fewer parameters than the two temporal models, but also exhibits a higher root error. Introducing temporal context provides a clear improvement in global localization. Despite relying on different temporal mechanisms, the TCN and Transformer have similar model sizes and achieve comparable World MPJPE. The TCN obtains the lowest root error, whereas the Transformer provides a slightly lower reprojection error.

These results reinforce the observation that exploiting temporal information is more important in our setting than the specific choice between convolutional and attention-based temporal modeling.

\begin{table}[t]
    \centering
    \caption{
    Comparison of the frame-wise and temporal architectures. The number of trainable parameters is reported together with the main localization metrics.
    }
    \label{tab:supp_architecture}
    \resizebox{\columnwidth}{!}{%
        \begin{tabular}{lrrr}
\toprule
Model & Root error $\downarrow$ (cm) & World MPJPE $\downarrow$ (cm) & Params. (M) \\
\midrule
MLP & 13.63 & 15.76 & 0.194 \\
TCN (41 f) & \textbf{10.12} & \textbf{13.20} & 1.278 \\
Transformer (41 f) & 11.04 & 13.21 & 1.252 \\
\bottomrule
\end{tabular}

    }
\end{table}

\subsection{Extended Input Ablation}
\label{sec:supp_input_ablation}

The main paper reports the input ablation using the TCN. Table~\ref{tab:supp_input_ablation} provides the corresponding results for both the TCN and Transformer.

Overall, the two architectures exhibit similar trends. Removing image-space pose and bounding-box cues produces the strongest degradation for both models. Camera information is also important, particularly for the Transformer. Removing the explicit ray--ground intersection deteriorates performance for both temporal backbones, supporting its use as an informative geometric cue in addition to its role in constructing the initial root estimate.

The relative 3D pose provides complementary information about the player's body configuration, while the valid-joint mask improves robustness to incomplete or unreliable observations. The consistency of these trends across two different temporal architectures suggests that the observed gains primarily originate from the information contained in the features rather than from architecture-specific behavior.

\begin{table*}[t]
    \centering
    \caption{
    Extended input ablation for the TCN and Transformer residual models. $\Delta$ Root denotes the increase in root error with respect to the corresponding full model.
    }
    \label{tab:supp_input_ablation}
    \resizebox{\textwidth}{!}{%
        \begin{tabular}{llrrrr}
\toprule
Model & Input configuration & Root $\downarrow$ (cm) & $\Delta$ Root (cm) & World MPJPE $\downarrow$ (cm) & Reproj. $\downarrow$ (px) \\
\midrule
TCN & Full & 10.12 & 0.00 & 13.20 & 3.49 \\
TCN & w/o valid-joint mask $\mathbf{m}_{i,t}$ & 10.75 & 0.63 & 13.53 & 3.41 \\
TCN & w/o ground intersection $\mathbf{q}^{w}_{i,t}$ & 11.17 & 1.05 & 13.87 & 3.43 \\
TCN & w/o relative 3D pose $\mathbf{f}^{3D}$ & 12.40 & 2.28 & 14.93 & 3.70 \\
TCN & w/o 2D pose and box cues $\mathbf{f}^{2D},\mathbf{f}^{\mathrm{box}}$ & 15.93 & 5.81 & 18.20 & 5.54 \\
TCN & w/o camera descriptor $\mathbf{f}^{cam}$ & 13.02 & 2.90 & 15.39 & 3.50 \\
Transformer & Full & 11.04 & 0.00 & 13.21 & 3.43 \\
Transformer & w/o valid-joint mask $\mathbf{m}_{i,t}$ & 13.42 & 2.39 & 15.03 & 3.69 \\
Transformer & w/o ground intersection $\mathbf{q}^{w}_{i,t}$ & 12.02 & 0.98 & 13.91 & 3.50 \\
Transformer & w/o relative 3D pose $\mathbf{f}^{3D}$ & 12.28 & 1.24 & 14.34 & 3.68 \\
Transformer & w/o 2D pose and box cues $\mathbf{f}^{2D},\mathbf{f}^{\mathrm{box}}$ & 15.81 & 4.77 & 17.41 & 5.20 \\
Transformer & w/o camera descriptor $\mathbf{f}^{cam}$ & 15.46 & 4.42 & 17.25 & 4.11 \\
\bottomrule
\end{tabular}
    }
\end{table*}

\subsection{Temporal Consistency}
\label{sec:supp_temporal_consistency}

In addition to positional accuracy, we evaluate the consistency of the predicted root trajectories through first- and second-order temporal differences.

Table~\ref{tab:supp_temporal_consistency} reports root-position, velocity, and acceleration errors for the frame-wise MLP and both temporal models. The MLP predicts each frame independently and consequently exhibits larger temporal errors. Both the TCN and Transformer considerably reduce velocity and acceleration errors while simultaneously improving root localization.

The TCN achieves the lowest root-position and velocity errors, whereas the Transformer obtains a marginally lower acceleration error. The proximity of their results again indicates that most of the gain comes from temporal context itself rather than from a specific temporal architecture.

\begin{table}[t]
    \centering
    \caption{
    Temporal consistency of the predicted root trajectories. Lower values indicate more accurate root position, velocity, and acceleration.
    }
    \label{tab:supp_temporal_consistency}
    \resizebox{\columnwidth}{!}{%
        \begin{tabular}{lrrr}
\toprule
Model & Root error $\downarrow$ (cm) & Root velocity error $\downarrow$ (cm s$^{-1}$) & Root acceleration error $\downarrow$ (m s$^{-2}$) \\
\midrule
MLP & 13.63 [12.50, 15.08] & 131.9 [124.1, 142.2] & 48.3 [45.6, 51.7] \\
TCN (41 f) & \textbf{10.12 [9.20, 11.33]} & \textbf{100.6 [95.3, 107.6]} & 39.6 [37.7, 42.2] \\
Transformer (41 f) & 11.04 [9.96, 12.42] & 101.9 [95.7, 109.8] & \textbf{39.0 [36.7, 42.1]} \\
\bottomrule
\end{tabular}

    }
\end{table}

\subsection{Temporal Gain Across Motion Regimes}
\label{sec:supp_temporal_regimes}

\begin{figure}[t]
    \centering
    \includegraphics[width=\columnwidth]{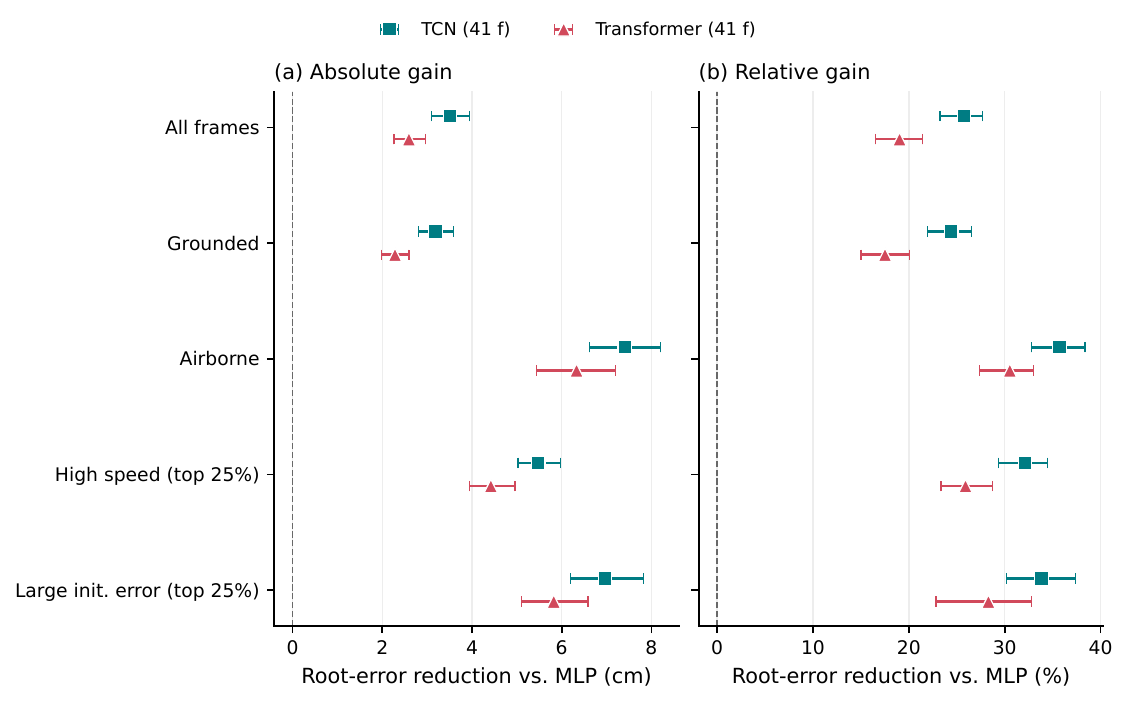}
    \caption{
    Relative improvement of temporal models over the frame-wise MLP across different motion regimes.
    }
    \label{fig:supp_temporal_regimes}
\end{figure}

Figure~\ref{fig:supp_temporal_regimes} analyzes the benefit of temporal modeling under different motion conditions. The improvement is reported relative to the frame-wise MLP.

Temporal modeling is beneficial across all evaluated regimes, but the improvement is larger in difficult situations. In particular, the TCN provides larger relative gains for airborne players, high-speed motion, and frames associated with large geometry-initialization errors than for grounded configurations.

This observation suggests that temporal context becomes particularly useful when instantaneous geometric and image-space observations are ambiguous. Neighboring frames can then provide information about the player's trajectory that is unavailable from a single frame.

\subsection{Sensitivity to Image-Space Position}
\label{sec:supp_image_position}

\begin{figure}[t]
    \centering
    \includegraphics[width=\columnwidth]{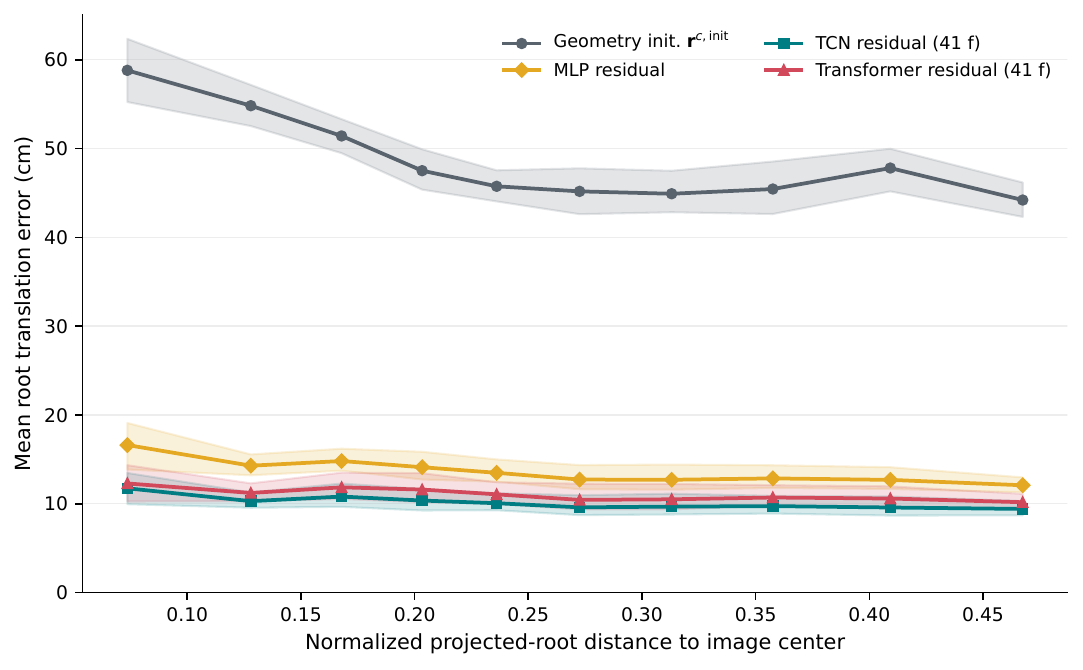}
    \caption{
    Root error as a function of player distance from the image center.
    }
    \label{fig:supp_image_center}
\end{figure}

Figure~\ref{fig:supp_image_center} reports root error as a function of the distance between the player and the image center.

No severe degradation is observed as players move away from the center of the broadcast image. This indicates that the model does not rely exclusively on a narrow image region and can exploit the camera and geometric descriptors to accommodate different viewing configurations.

\subsection{Sensitivity to Player Image Scale}
\label{sec:supp_bbox}

\begin{figure}[t]
    \centering
    \includegraphics[width=\columnwidth]{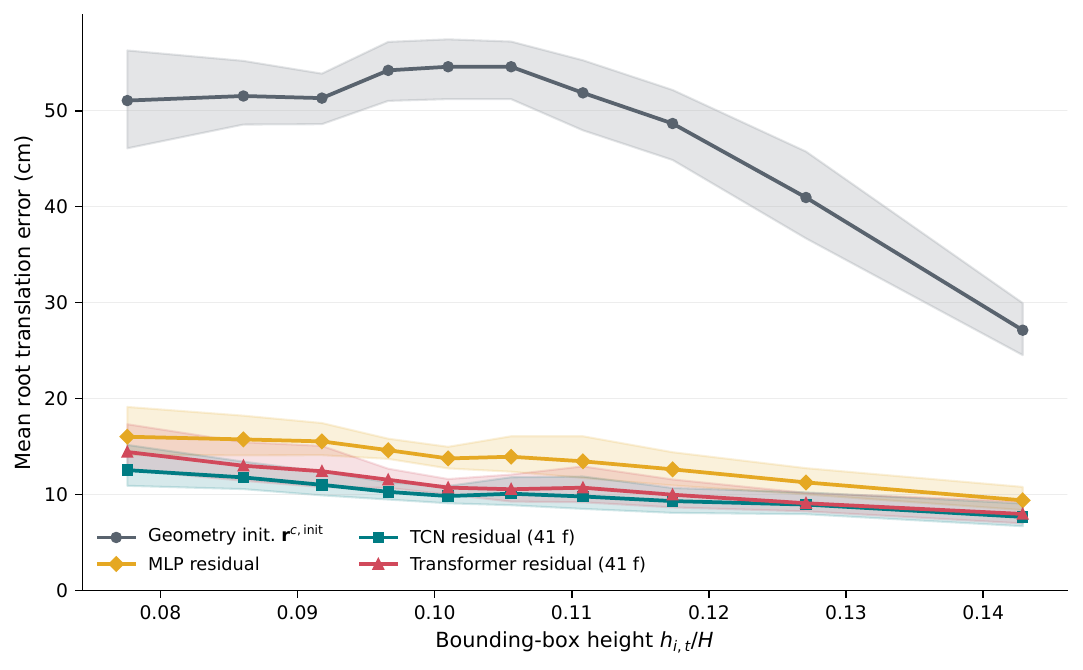}
    \caption{
    Root error as a function of player bounding-box height in the image.
    }
    \label{fig:supp_bbox_height}
\end{figure}

Figure~\ref{fig:supp_bbox_height} evaluates root localization as a function of player bounding-box height.

Localization is more difficult for players occupying fewer pixels in the image, as expected in broadcast footage. Smaller players provide less accurate image-space body observations and weaker perspective cues. Performance progressively improves as the apparent player size increases.

This result highlights image resolution as an important source of uncertainty for world-grounded reconstruction, particularly for players located far from the broadcast camera.

\subsection{Sensitivity to Player Motion}
\label{sec:supp_speed}

\begin{figure}[t]
    \centering
    \includegraphics[width=\columnwidth]{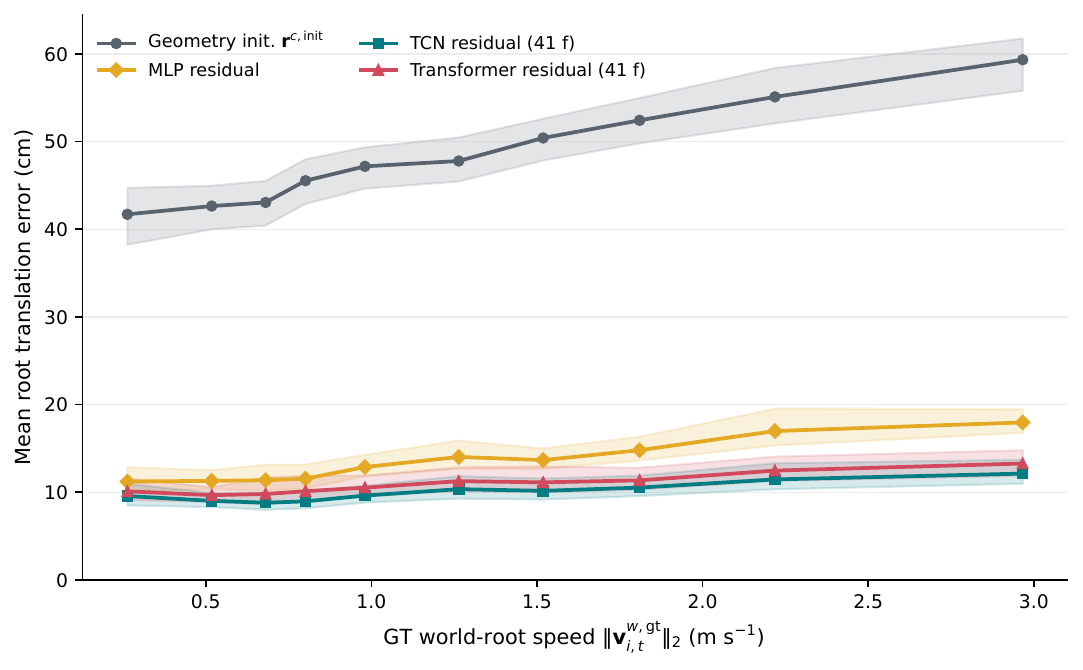}
    \caption{
    Root error as a function of player speed.
    }
    \label{fig:supp_speed}
\end{figure}

Figure~\ref{fig:supp_speed} analyzes root error as a function of player speed.

Localization error increases for the fastest motions. High-speed actions are associated with larger frame-to-frame displacement, stronger motion blur, rapidly changing body configurations, and potentially less reliable instantaneous keypoints. Nevertheless, the temporal models remain more accurate than the frame-wise baseline in this regime, consistent with the results of Sec.~\ref{sec:supp_temporal_regimes}.

\subsection{Root Error Along Camera Axes}
\label{sec:supp_camera_axes}

\begin{figure}[t]
    \centering
    \includegraphics[width=\columnwidth]{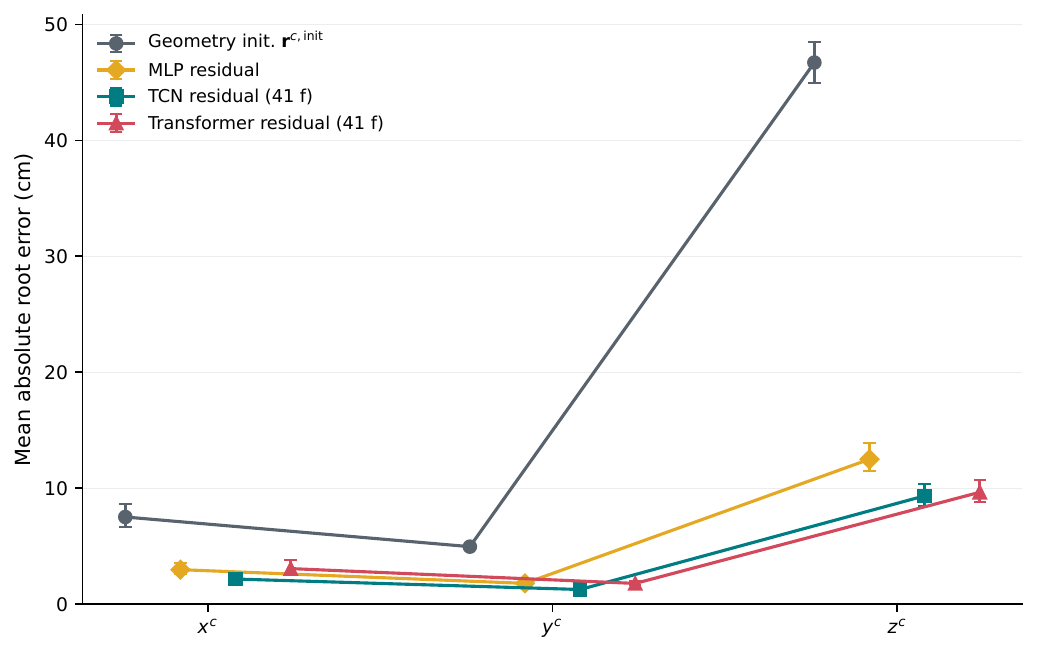}
    \caption{
    Root localization error decomposed along the camera coordinate axes.
    }
    \label{fig:supp_camera_axes}
\end{figure}

To better understand the source of global localization error, Fig.~\ref{fig:supp_camera_axes} decomposes the root error along the camera coordinate axes.

The geometry-based initialization is substantially more accurate in the lateral image-plane directions than along camera depth. Temporal residual refinement strongly reduces all three components, but depth remains the dominant source of localization error.

This behavior is consistent with the inherent ambiguity of monocular reconstruction: small image-space errors can correspond to large metric displacements along the viewing direction. The result further motivates the use of temporal and scene-geometric cues for global player localization.

\subsection{Pelvis Height Analysis}
\label{sec:supp_pelvis_height}

\begin{figure}[t]
    \centering
    \includegraphics[width=\columnwidth]{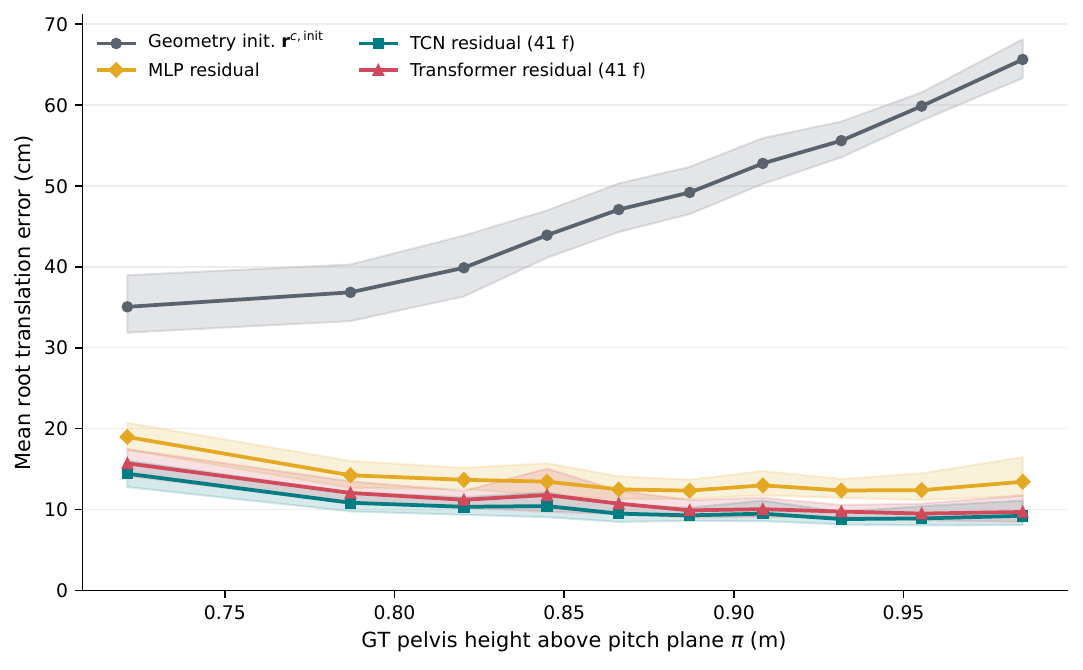}
    \caption{
    Root error as a function of pelvis height above the playing surface.
    }
    \label{fig:supp_pelvis_height}
\end{figure}

For completeness, Fig.~\ref{fig:supp_pelvis_height} reports root error as a function of pelvis height above the pitch.

Unlike lower-foot clearance, pelvis height does not provide a direct indicator of ground contact. It varies with player morphology and body configuration, including flexion and extension during otherwise grounded movements. Consequently, its relationship with localization error is less direct than the airborne analysis presented in the main paper.

We therefore use lower-foot clearance as the primary measure for analyzing violations of the ground-contact assumption and provide pelvis-height results only as a complementary diagnostic.

%

\subsection{Measured Training and Evaluation Throughput.}
\label{sec:computational_efficiency}
\begin{figure}[t]
    \centering
    \includegraphics[width=\columnwidth]
    {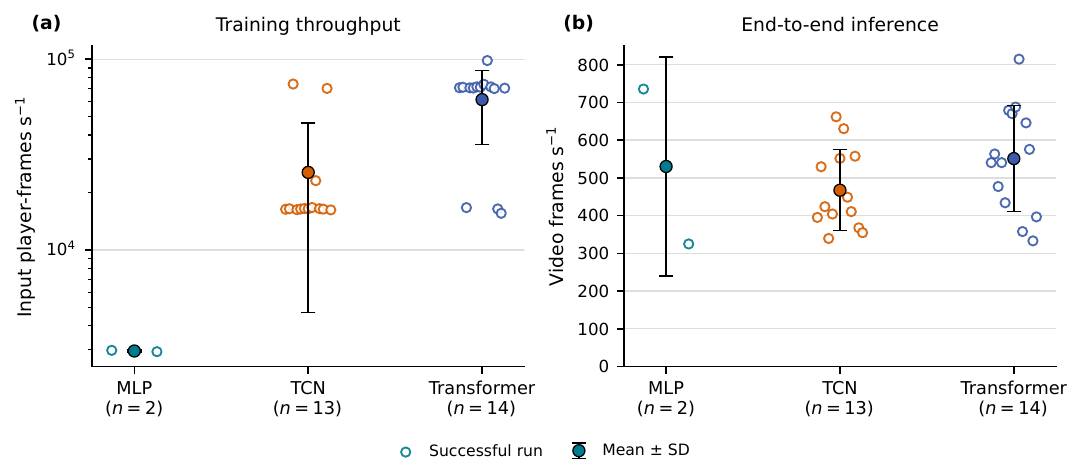}
    \caption{
        Computational throughput of the MLP, TCN, and Transformer architectures on an NVIDIA L40S GPU.
        \textbf{(a)} Training throughput in input player-frames per second.
        \textbf{(b)} End-to-end inference throughput in video frames per second.
        Each hollow marker represents one successfully completed run, while filled markers and error bars indicate the mean and standard deviation across runs.
        Training throughput counts the player-frame tokens processed by the optimization loop, including overlapping temporal windows.
    }
    \label{fig:runtime_throughput}
\end{figure}

Runtime was measured on a single NVIDIA L40S GPU (46,068~MiB visible memory) using PyTorch~2.6 and CUDA~12.4. We considered all 29 successfully completed full-data runs from the unseen-match protocol: 2 MLP, 13 TCN, and 14 Transformer runs. As shown in Fig.~\ref{fig:runtime_throughput}, mean training throughput was $(2.95\pm0.03)\times10^3$, $(25.5\pm20.8)\times10^3$, and $(61.4\pm25.5)\times10^3$ input player-frames~s$^{-1}$ for the MLP, TCN, and Transformer, respectively. End-to-end batched evaluation, including data loading, temporal-window aggregation where applicable, metric computation, and prediction serialization, reached $530\pm290$, $467\pm107$, and $551\pm141$ video frames~s$^{-1}$. Values denote mean $\pm$ standard deviation across successful configurations; the dispersion therefore reflects variations in input features and temporal context rather than repeated-run confidence intervals. These measurements exclude upstream player tracking, camera calibration, and SAM3DBody inference.

\end{document}